\pdfoutput=1

\documentclass[11pt,twoside]{article}
\usepackage[letterpaper,inner=3.5cm,outer=2.5cm,top=2.5cm,bottom=2.5cm,pdftex]{geometry}
\usepackage[T1]{fontenc}
\usepackage[utf8]{inputenc}
\usepackage{lmodern}
\usepackage{authblk}
\usepackage[pdftex]{graphicx}
\usepackage[pdftex]{hyperref}
\usepackage{fancyhdr}

\usepackage[utf8]{inputenc} 
\usepackage[T1]{fontenc}    
\usepackage{hyperref}       
\usepackage{url}            
\usepackage{booktabs}       
\usepackage{amsfonts}       
\usepackage{nicefrac}       
\usepackage{microtype}      

\usepackage{subfigure}
\usepackage{braket}

\usepackage{amsmath}
\usepackage{amsthm}
\usepackage{mathtools}
\usepackage{lmodern}

\usepackage[square,comma,numbers]{natbib}
\usepackage{amsmath}
\usepackage{dsfont}
\usepackage{mathtools}
\usepackage{algorithm}
\usepackage{algpseudocode}
\usepackage{enumitem}

\newcommand{\argmin}{\mathrm{argmin}}

\renewcommand{\vec}[1]{{#1}}

\newcommand{\norm}[1]{\left\|\,#1\,\right\|}   

\newcommand{\RNum}[1]{\uppercase\expandafter{\romannumeral #1\relax}}
\newcommand{\complex}{{\mathbb C}}

\newcommand{\Rset}{{\mathbb R}}

\newcommand{\BigPolyOh}[1]{\tilde{O}\left({#1}\right)}

\newcommand{\mute}[1]{}
\newcommand{\wrong}[1]{}
\newcommand{\research}[1]{}
\newcommand{\but}[1]{}

\newcommand{\half}{\frac{1}{2}}

\newcommand{\diverg}{\mathrm{div}}
\newcommand{\spc}{{\;\;\;}}

\newcommand{\summ}[1]{\sum_{{#1}=1}^m}
\newcommand{\fnal}[1]{{#1}}

\newcommand{\Fspace}{\mathcal{F}}

\newcommand{\inX}{\mathcal{X}}

\newcommand{\Hnorm}[1]{||#1||_{\mathcal{H}}}

\newcommand{\innerPar}[3]{\langle #1 , #2 \rangle_{{#3}}}

\newcommand{\Fnorm}[1]{||#1||_{\mathcal{F}}}
\newcommand{\FnormSq}[1]{||#1||^2_{\mathcal{F}}}
\newcommand{\innerF}[2]{\langle #1 , #2 \rangle_{\mathcal{F}}}

\newcommand{\featAtT}[1]{{\mathcal{T}}}

\newcommand{\rhoA}{K^{\dagger}}
\newcommand{\rhoB}{L}
\newcommand{\rhoC}{K}

\newcommand*{\mydprime}{^{\prime\prime}\mkern-1.2mu}
\newcommand*{\mytrprime}{^{\prime\prime\prime}\mkern-1.2mu}

\newcommand{\rhoY}{\rho\mydprime}
\newcommand{\rhoZ}{\rho\mytrprime}

\title{Quantum Semi-Supervised Kernel Learning}

\author{Seyran Saeedi\thanks{e-mail: saeedis@vcu.edu}}
\author{Aliakbar Panahi\thanks{e-mail: panahia@vcu.edu}}
\author{Tom Arodz\thanks{Corresponding author. e-mail: tarodz@vcu.edu}}
\affil{\mbox{Department of Computer Science}, \mbox{Virginia Commonwealth University} \mbox{Richmond, VA, USA}}
\date{}

\begin{document}
	\sloppy
	\maketitle

\begin{abstract}
Quantum computing leverages quantum effects to build algorithms that are faster then their classical variants. In machine learning, for a given model architecture, the speed of training the model is typically determined by the size of the training dataset. Thus, quantum machine learning methods have the potential to facilitate learning using extremely large datasets. While the availability of data for training machine learning models is steadily increasing, oftentimes it is much easier to collect feature vectors that to obtain the corresponding labels. One of the approaches for addressing this issue is to use semi-supervised learning, which leverages not only the labeled samples, but also unlabeled feature vectors. Here, we present a quantum machine learning algorithm for training Semi-Supervised Kernel Support Vector Machines. The algorithm uses recent advances in quantum sample-based Hamiltonian simulation to extend the existing Quantum LS-SVM algorithm to handle the semi-supervised term in the loss. Through a theoretical study of the algorithm's computational complexity, we show that it maintains the same speedup as the fully-supervised Quantum LS-SVM.
\end{abstract}

\section{Introduction}

Building computing machines that can exploit quantum effects has lead to faster algorithms in many fields, including number theory \citep{shor1994algorithms}, database search \citep{grover1996fast}, and linear algebra  \citep{harrow2009quantum}. One of the fields where quantum computing has potential to offer substantial gains is machine learning. Quantum algorithms for training machine learning models have expanded into an active field of research \citep{BWPR17,dunjko2018machine,SchF18,arunachalam2017guest}. The study of quantum algorithms has also led to discovery of more efficient classical methods \citep{stoudenmire2016supervised,tang2019quantum,panahi2019word2ket,panahi2021shapeshifter}.

Quantum machine learning aims to provide quantum algorithms that can speed up training of the machine learning models on large datasets. 
An early example of this approach is Quantum Least-Squares Support Vector Machine (LS-SVM) \citep{rebentrost2014quantum}, which achieves exponential speedup compared to classical LS-SVM algorithm. Quantum LS-SVM uses quadratic least-squares loss and squared-$L_2$ regularizer, and the optimization problem can be solved using the seminal HHL \citep{harrow2009quantum} algorithm for solving quantum linear systems of equations. More recently, a quantum training algorithm that solves a maximin problem resulting from a maximum – not average -- loss over the training set has been proposed \citep{li2019sublinear}. A sparse SVM method that uses quantum linear programming solvers and achieves quadratic speedup for certain families of input problems  has also been introduced recently \citep{arodz2019quantum}.  Most recently, a quantum algorithm that uses efficient quantum approximations of inner products in the kernel space within the SVM-perf frameworks has also been introduced  \citep{allcock2020quantum}, achieving complexity linear in the number of samples. 
All these variants of quantum SVM are aimed at supervised problems, where each training sample is accompanied by its true class label. While progress has been made in quantum algorithms for supervised learning, it has been recently advocated that the focus should shift to unsupervised and semi-supervised setting \citep{perdomo2018opportunities}.

In many domains, the most laborious part of assembling a training set is collecting sample labels. Thus, in many scenarios, in addition to the labeled training set, we have access to many more feature vectors with missing labels. One way of utilizing these additional data points to improve the classification model is through semi-supervised learning. In semi-supervised learning, we are given $m$ observations $x_1,...,x_m \in \inX$ drawn from the marginal distribution $p(x)$ over some input space $\inX$. The first $l$ ($l \ll m$) data points come with labels $y_1,...,y_l$ drawn from conditional distribution $p(y|x)$. Semi-supervised learning algorithms exploit the underlying marginal distribution of the data, $p(x)$, estimated using both labeled and unlabeled data points, to improve classification accuracy on unseen samples. 

Here, we introduce a quantum algorithm for semi-supervised training of a kernel support vector machine classification model. We start with the existing Quantum LS-SVM \citep{rebentrost2014quantum}, and use techniques from sample-based Hamiltonian simulation \citep{kimmel2017hamiltonian} to add a semi-supervised term based on Laplacian SVM \citep{melacci2011laplacian}. As is standard in quantum machine learning \citep{li2019sublinear}, the algorithm accesses training points and the adjacency matrix of the graph connecting samples via a quantum oracle. We show that, with respect to the oracle, the proposed algorithm achieves the same quantum speedup as LS-SVM, that is, adding the semi-supervised term does not lead to increased computational complexity.

\section{Preliminaries}
In this section, we setup the background that we need for describing quantum semi-supervised algorithm. We first describe briefly the semi-supervised kernel machines and show that training a semi-supervised LS-SVM problem can be reformulated as a system of linear equations. Second, we describe the quantum subroutines used in the Quantum LS-SVM \citep{rebentrost2014quantum}, which  minimizes quadratic loss and quadratic regularizer $\sum_{i=1}^m (y_i-\beta^T x_i)^2 + \lambda \lVert \beta \rVert_2^2$ over the training set $\left\{(x_i,y_i)\right\}$ and with standard techniques can be re-written as a system of linear equations that can be solved using quantum linear algebra approaches. Third, we describe the basic quantum subroutines we utilize such as the HHL algorithm \citep{harrow2009quantum} and the LMR protocol \citep{lloyd2014quantum}.

\subsection{Semi-Supervised Kernel Machines}
\label{secRKHS}

\paragraph{Reproducing Kernel Hilbert Spaces.}

Binary classification models take the form of a function $f: \inX \rightarrow \Rset$. Consider functions from $L_2(\inX)$, the space of all square-integrable functions  $\inX \rightarrow \Rset$. In $L_2$, closeness in norm does not imply everywhere pointwise closeness of two functions. Difference between $f(x)$ and $g(x)$ can be arbitrary large for $x \in S \subset \inX$ even if $\Hnorm{f-g}=0$, as long as $S$ has measure of zero. 

One large class of functions that does not exhibit this problem is the Reproducing Kernel Hilbert Space (RKHS) of functions. Its construction is based on Dirac evaluation functionals in the following way. 
For a Hilbert space $\Fspace$ of functions $f: \inX \rightarrow \Rset$, consider a family of Dirac evaluation functionals $F$ consisting of mappings $F_t: \Fspace \rightarrow \Rset$ parameterized by a specific argument $t \in \inX$ and, for any $f \in \Fspace$, returning the value of $f$ evaluated at $t$; that is, $\fnal{F}_t[h]=f(t)$. Dirac evaluation functionals with the same parameter $t$ are linear, that is, $\alpha F_t[f]+\beta F_t[g] = F_t[\alpha f+ \beta g] = \alpha f(t)+ \beta g(t)$. 
Any linear functional $\fnal{F}: \Fspace \rightarrow \Rset$ is bounded if 
$\exists M \in \Rset : \spc \forall f \in \Fspace \spc |\fnal{F}[f]| \leq M \Hnorm{f}$.
If in the space $\Fspace$ all Dirac evaluation functionals for all $t\in X$ are bounded, then, for any $f,g \in \Fspace$, small $\Fnorm{f-g}$ implies small $|f(t)-g(t)|$, everywhere on $X$; closeness of functions in norm implies pointwise closeness. 
Riesz representation theorem guarantees that for each bounded evaluation functional $\fnal{F}_t$, there exists a unique function $K_t \in \Fspace$ such that $\fnal{F}_t[h]=fh(t)=\innerF{K_t}{f}$.
Function $K_t: \inX \rightarrow \Rset$ is called the {\em representer} of $t \in \inX$. Any space of functions in which every Dirac evaluation operator $\fnal{F}_t$ is bounded, and thus has a corresponding representer $K_t$, is called a {\em Reproducing Kernel Hilbert Space}.

In RKHS, by symmetry of inner product, $K_t(s)=\innerF{K_t}{K_s}=K_s(t)$. We can thus define a function $K:\inX \times \inX \rightarrow \Rset$ with values $K(s,t)=K_s(t)$, such that 
$\forall x,y \in \inX \spc \exists K_x, K_y \in \Fspace: \spc \innerF{K_x}{K_y}=K(x,y)$.
Any symmetric and positive definite function $K:\inX \times \inX \rightarrow \Rset$ (that is, a function that fulfills the Mercer condition,  $\int_{\inX} \int_{\inX}  c(x) K(x,y) c(y) dx dy \geq 0 \spc \forall c \in \Fspace$ ) is called a {\em reproducing kernel}. A reproducing kernel gives rise to functions $K_t: \inX \rightarrow \Rset$ defined by fixing $t$ and defining $K_t(x)=K(t,x)$. Given the input space $\inX$, we can construct an inner product space that is an RKHS in the following way. The RKHS space will consist of  all possible finite linear combinations of representers, that is, of all functions of the form  $f=\sum_{j=1}^n c_j K_{t_j}$ for some finite $n$, some representers $\left\{K_{t_j}\right\}$ and some weights $\left\{c_j\right\}$, with $j=1,...,n$. Given two functions $f=\sum_{j=1}^n c_j K_{t_j}$ and $g=\sum_{i=1}^{n'} c'_i K_{t'_i}$ for some finite $n$ and $n'$, representer sets $\left\{K_{t_j}\right\}$ and $\left\{K_{t'_i}\right\}$, and their linear combination weights  $\left\{c_j\right\}$ and $\left\{c'_i\right\}$, the inner product in RKHS is defined through the reproducing kernel $K$ as $\innerF{f}{g}=\sum_{j=1}^n \sum_{i=1}^{n'} c_j c'_i K(t_j,t'_i)$. The Moore-Aronszajn theorem states that the space defined this way can be completed, the resulting RKHS space is unique, and $K$ is the reproducing kernel in that space. Kernel method, including kernel SVM, involve functions from RKHS as predictors $f(x) \rightarrow y$, and rely on  expressing the inner product via the reproducing kernel $\innerF{K_x}{K_y}=K(x,y)$ to have the efficient way of working with functions from RKHS.

\paragraph{Discrete-topology Gradients of Functions in RKHS.}
\label{seclap}

Kernel methods including SVMs are regularized by penalizing for high local variability of the predictor function $f$. This makes the predictions more stable to small changes in input, and leads to improved generalization capabilities of the trained classifier. A simple way of measuring local variability of a function $\inX \rightarrow \Rset$ is through its gradient; for example, functions with Lipschitz-continuous gradient cannot change too rapidly. In manifold learning \citep{belkin2006manifold}, instead of or in addition to analyzing gradient using the topology of $\inX$, some other, more relevant topology is used. In particular, in Laplacian SVM \citep{melacci2011laplacian} and related semi-supervised learning methods, a discrete topology of a graph $G$ connecting training points in $\inX$ is used. In training semi-supervised models, we assume that we have information on the similarity of the training samples, in a form of a graph $G$ that uses $n$ edges to connect similar training samples, represented as $m$ vertices. The graph includes vertices with and without labels, providing a consistent view of the underlying manifold on which all available samples lie. For each sample $i$, by $d_i$ we denote the number of its neighbors in graph $G$, that is, the degree of vertex $i$ is $d_i$. The graph connecting samples can be available a priori, or can be constructed as a pre-processing step by adding an edge between each vertex and its $d_i$ nearest neighbors, or connecting a vertex to all neighbors with similarity above a user-defined threshold. In general, degree in graph $G$ may differ for different vertices. Ultimately, in training a semi-supervised classifier, we assume that the graph $G$ is given on input, and it has all training samples, including those without class labels, as vertices. 

For a given undirected graph $G$ with a set of  $m$ vertices, $V$, and a set of $n$ edges, $E$, let us define a Hilbert space $\Fspace_V$ of functions $f: V \rightarrow \Rset$ with inner product $\innerPar{f}{g}{V}=\sum_{v \in V} f(v)g(v)$, and a Hilbert space $\Fspace_E$ of functions $\psi: E \rightarrow \Rset$ with inner product $\innerPar{\psi}{\phi}{E}=\sum_{e \in E} \psi(e)\phi(e)=\sum_{u \sim v} \psi([u,v])\phi([u,v])$, where $u \sim v$ represents an undirected edge connecting vertices $u$ and $v$ in the input graph $G$.
Given these two spaces, we define a linear operator $\nabla: \Fspace_V \rightarrow \Fspace_E$ such that
\begin{align*}
\nabla f([u,v])=\sqrt{G_{u,v}} f(u) - \sqrt{G_{u,v}} f(v) = -\nabla f ([v,u]),
\end{align*}
where $G_{u,v}$ represents the weight of the edge $u \sim v$; in semi-supervised learning explored here, all weights are uniform, $G_{u,v}=1$ for any edge $u \sim v$. We set $G_{u,v}=0$ if $u$ and $v$ are not connected by an edge.

The operator $\nabla$ can be seen as a discrete counterpart to the gradient of a function -- given a function $f$ with values for training samples or, equivalently, the corresponding vertices, for a given sample/vertex $u$ in the domain of $f$, $\nabla$ over different samples/vertices $v$ gives us a set of values showing the change of $f$ over all directions from $u$, that is, all edges incident to it.
Based on the operator $\nabla$, we can define the equivalent of divergence over the graph topology, a linear operator $\diverg: \Fspace_E \rightarrow \Fspace_V$ such that $-\diverg$ is the adjoint of $\nabla$, that is, $\innerPar{\nabla[f]}{\psi}{E}=\innerPar{f}{-\diverg[\psi]}{V}$. Finally, we define {\em graph Laplacian}, a linear operator $\Delta: \Fspace_V \rightarrow \Fspace_V$
\begin{align*}
\Delta[f]&=-\frac{1}{2} \diverg[\nabla[f]] =D_v f(v)-\sum_{u \sim v}G_{u,v} f(u).
\end{align*}
The graph Laplacian operator $\Delta$ is self-adjoint and positive semi-definite, and the squared norm of the graph gradient can be captured through it as 
\begin{align*}
\frac{1}{2} || \nabla f ||^2_E&=\innerPar{\Delta[f]}{f}{V} = \frac{1}{2}\sum_{u \sim v} G_{u,v}  (\bar{f}_u - \bar{f}_v )^2=\bar{f}^T L \bar{f}.
\end{align*}
In the above definition, we used $\bar{f}$ to denote an $m$-dimensional vector indexed by the graph vertices and consisting of values of $f$ for the samples corresponding to the vertices; that is, $\bar{f}_u = f(u)$. Then, $\Delta[f]$ can be expressed as $\Delta[f]=L \bar{f}$, a multiplication of vector of function values $\bar{f}$ by the {\em combinatorial graph Laplacian matrix} $L$ such that ${L}[i,i]=D_i$ and ${L}[i,j]= - G_{i,j}$. 

In semi-supervised learning, we use the norm $|| \nabla f ||^2_E$ as another regularization factor. While the regularization factor $||f||^2$ used in the regular SVM promotes smoothness of the predictive model $f$ in the original topology, the additional Laplacian-based term promotes smoothness of $f$ over the graph topology. Vertices that may be relatively close in the original topology may be distant in terms of graph topology, providing for a more appropriate regularization. Further, the graph topology is often better at capturing the manifold on which the input samples reside \citep{belkin2006manifold}. The graph Laplacian is based on the graph connecting the samples, and the degrees of the vertices may span a wide range of value, for example if a similarity threshold in the input space is used to define connectivity of samples and the density of samples varies in the input space. Therefore, in practical usage, instead of the combinatorial graph Laplacian, one often employs its degree-normalized version, with ${L}[i,i]=1$ and ${L}[i,j]= - G_{i,j} / \sqrt{D_i D_j}$. 

\paragraph{Semi-Supervised Least-Squares Kernel Support Vector Machines.}

Our goal is to find predictors $f(x):\inX \rightarrow \Rset$ that are functions from a RKHS $\Fspace$ defined by a kernel $K$.  In Semi-Supervised LS-SVMs in RKHS, we are looking for a function $f \in \Fspace$ that minimizes
\begin{align}
\label{eqlossSVM}
\argmin_{f\in \Fspace}   \frac{\gamma}{2}\sum_{i=1}^{l} \left( y_{i} -  f(x_i)  \right)^2
+ \half \FnormSq{f} + \half || \nabla f ||^2_E, 
\end{align}
where $l \leq m$ is the number of training points with labels (these are grouped at the beginning of the training set of size $m$),  and $\gamma$ is a user-defined constant allowing for adjusting the regularization strength. The first two terms are the same as in the standard LS-SVM -- they include a least-squares penalty  for predictions, $\left(y_{i}  f(x_i) \right)^2$, and an $L_2$ regularization of predictions in the kernel space, $\FnormSq{f}$.  Compared to standard LS-SVM, the objective function contains additional, third term: $|| \nabla f ||^2_E$; as indicated above, this term corresponds to a quadratic term involving the graph Laplacian of a graph connecting all available labeled and unlabeled samples. It promotes a smoothness of the predictive function $f$ over the graph, that is, if promotes solutions $f$ in which similar training samples $x$ and $x'$ have similar predictions $f(x)$ and $f(x')$ if the two samples are connected by an edge in the graph.
 
The Representer Theorem \citep{scholkopf2001generalized} states that if $\Fspace$ is RKHS defined by kernel $K:\inX \times \inX \rightarrow \Rset$, then the solution minimizing the problem above is achieved for a function $f$ that uses only the representers of the training points, that is, a function of the form
$f(x)= \summ{j}\alpha_j K_{x_j}(x)=\summ{j}\alpha_j K(x_j,x)$; there is nothing to be gained by utilizing representers of points not in the training set.
Here, coefficients $\alpha_j$ can assume arbitrary real values; in some equivalent specifications of SVM, we alternatively see terms $y_j \alpha_j$ instead, with $\alpha_j\geq 0$. Also, as advocated for kernel SVMs \citep{steinwart2011training}, we do not use any offset term. By using the definitions from the paragraphs above, we can translate the problem
\begin{align*}
\argmin_{f\in \Fspace} \frac{\gamma}{2}\sum_{i=1}^{l} \left( 1 - 2y_{i} f(x_i)  +  f(x_i)^2 \right) + \half \FnormSq{f} + \half || \nabla f ||^2_E
\end{align*}
into a  quadratic optimization problem over finite, real vectors $\vec{\alpha}$
\begin{align*}
\argmin_{\alpha}  \sum_{i=1}^{l}  \left(-\gamma y_{i}\sum_{j=1}^{m} \alpha_{j} K[i, j] +  \frac{\gamma}{2}\left( \sum_{j=1}^{m} \alpha_{j} K[i, j] \right)^2 \right)
+\frac{1}{2} \vec{\alpha}^{T} K \vec{\alpha}+\frac{1}{2} \vec{\bar{f}} L \vec{\bar{f}},
\end{align*}
where $\vec{\bar{f}}$ is the vector of function $f$ values at all training points; we have $\vec{\bar{f}}=K\vec{\alpha}$, or $f(x_i)=\sum_{j=1}^m \alpha_j K[i,j]$, since function $f$ is defined using representers $K_{\vec{x_i}}$ of all training points. The semi-supervised term, the squared norm of the graph gradient of $f$, $1/2 || \nabla f ||^2_E = \vec{\bar{f}}^TL\vec{\bar{f}}=\vec{\alpha}^TKLK\vec{\alpha}$, penalizes large changes of function $f$ over edges of graph $G$. 
In defining the kernel $K$ and the Laplacian $L$ and in the two regularization terms, we use all $m$ samples. On the other hand, in calculating the empirical quadratic loss, we only use the first $l$ samples.

To simplify notation, we introduce $y_i=0$ for the samples that lack labels, that is, for $l < i \leq m$, leading to $m$-dimensional vector $\vec{y}$. The quadratic problem can then be equivalently expressed in matrix form as 
\begin{align*}
\argmin_{\alpha} -\gamma \vec{\alpha}^T K\vec{y} + \frac{\gamma}{2} \vec{\alpha}^T K^T K \vec{\alpha}  +\frac{1}{2} \vec{\alpha}^{T} K \vec{\alpha} + \frac{1}{2} \vec{\alpha}^{T} K L K \vec{\alpha}.
\end{align*}

Since matrices $K$ and $L$ involved in this quadratic problem are positive semidefinite, the problem is convex, and the minimum can be obtained by equating the gradient to null, that is, by solving the following system of linear equations
\begin{align}
\label{eqsvmlineq}
\frac{1}{\gamma} K \vec{\alpha} +  K K \vec{\alpha}  + \frac{1}{\gamma}   K L K \vec{\alpha} =  K\vec{y}, 
\end{align}
where $\vec{y}=(y_1,...,y_m)^T$, $K$ is an $m \times m$ kernel matrix, $L$ is the an $m \times m$ graph Laplacian matrix, $\gamma$ is a hyperparameter and $\vec{\alpha}=(\alpha_1,...,\alpha_m)^T$ is the unknown vector of sample weights in the kernel SVM.

\subsection{Quantum Computing and Quantum LS-SVM} 
 
\paragraph{Quantum Linear Systems of Equations.}
Given an input matrix $A\in \complex^{m\times m}$ and a vector $b\in \complex^m$, the goal of linear system of equations problem is finding $x\in \complex^m$ such that $Ax=b$. When $A$ is Hermitian and full rank, the unique solution is $x=A^{-1}b$. If $A$ is not a full rank matrix then $A^{-1}$ is replaced by the Moore-Penrose pseudo-inverse. HHL algorithm introduced an analogous problem in quantum setting: assuming an efficient algorithm for preparing $b$ as a quantum state $b=\sum_{i=1}^m b_i\ket{i}$ using $\lceil \log m\rceil+1$ qubits, the algorithm applies quantum subroutines of phase estimation, controlled rotation, and inverse of phase estimation to obtain the state
\begin{equation*}
	\ket{x}=\frac{A^{-1}\ket{b}}{\norm{A^{-1}\ket{b}}}.
\end{equation*}    
Intuitively, HHL algorithm works as follows: if $A$ has spectral decomposition $A=\sum_{i=1}^m \lambda_i v_iv_i^T$ (where $\lambda_i$ and $v_i$ are corresponding eigenvalues and eigenstates of $A$), then $A^{-1}$ maps $\lambda_i v_i\mapsto \dfrac{1}{\lambda_i}v_i$. The vector $b$ also can be written as the linear combination of the $A$'s eigenvectors $v_i$ as $b=\sum_{i=1}^m \beta_i v_i$ (we are not required to compute $\beta_i$). Then $A^{-1}b=\sum_{i=1}^m \beta_i \dfrac{1}{\lambda_i} v_i$. In general $A$ and $A^{-1}$ are not unitary (unless all $A$'s eigenvalues have unit magnitude), therefore we are not able to apply $A^{-1}$ directly on $\ket{b}$. However, since $U=e^{iA}=\sum_{i=1}^m e^{i\lambda_i} v_iv_i^T$ is unitary and has the same eigenvectors as $A$ and $A^{-1}$, one can implement $U$ and powers of $U$ on a quantum computer by Hamiltonian simulation techniques; clearly for any expected speed-up, one need to enact $e^{iA}$ efficiently. The HHL algorithm uses the phase estimation subroutine to estimate an approximation of $\lambda_i$ up to a small error. The next step computes a conditional rotation on the approximated value of $\lambda_i$ and an auxiliary qubit $\ket{0}$ and outputs $\dfrac{1}{\lambda_i}\ket{0}+\sqrt{1-\dfrac{1}{\lambda_i^2}}\ket{1}$. The last step involves the inverse of phase estimation and quantum measurement for eliminating of garbage qubits and for returning the desired state $\ket{x}=A^{-1}\ket{b}=\sum_{i=1}^m \beta_i \dfrac{1}{\lambda_i} v_i$. The conditional rotation can be also implemented in a manner that does not result in the inverse of the eignevalues \citep{wiebe2012quantum}, to produce $\lambda_i\ket{0}+\sqrt{1-\lambda_i^2}\ket{1}$, leading to quantum matrix multiplication, $\ket{x}=\frac{A\ket{b}}{\norm{A\ket{b}}}$.

\paragraph{Quantum Estimation of the Kernel Matrix.}

Quantum LS-SVM uses density operator formalism and partial trace to represent the computation involving the kernel matrix. Given $m$ training samples, to obtain the corresponding $m \times m$ kernel matrix as a density matrix, quantum LS-SVM \citep{rebentrost2014quantum} relies on partial trace, and on  a quantum oracle that can convert, in superposition, each data point $\{x_i\}_{i=1}^m$, $x_i\in \mathbb{R}^p$ to a quantum state $\ket{x_i}=\frac{1}{\norm{x_i}}\sum_{k=1}^p (x_i)_k\ket{k}$, where $(x_i)_k$ refers to the $k$-th feature value in data point $x_i$ and assuming the oracle is given $\norm{x_i}$ and $y_i$. Vector of the labels is given in the same fashion as $\ket{y}= \frac{1}{\norm{y}}\sum_{i=1}^m y_i\ket{i}$. For preparation the normalized kernel matrix $K^\prime =\frac{1}{\operatorname{tr}(K)}K$ where $K=X^TX$, we need to prepare a quantum state combining all data points in quantum superposition $\ket{X}=\frac{1}{\sqrt{\sum_{i=1}^m \norm{x_i}^2}} \sum_{i=1}^m\ket{i}\otimes \norm{x_i}\ket{x_i}$. The normalized Kernel matrix is obtained by discarding the training set part of the quantum system, 
\begin{equation}
\begin{split}
K^\prime &= \operatorname{Tr}_2(\ket{X}\bra{X})=\frac{1}{\sum_{i=1}^m \norm{x_i}^2} \sum_{i,j=1}^m \norm{x_i}\norm{x_j}\braket{x_i|x_j}\ket{i}\bra{j}.
\end{split}	
\end{equation}
The approach used above to construct density matrix corresponding to linear kernel matrix can be extended to polynomial kernels \citep{rebentrost2014quantum}.

\paragraph{LMR Protocol for Density Operator Exponentiation.}
In HHL-based quantum machine learning algorithms that involve matrix inversion or matrix multiplication, including in the method proposed here, matrix $A$ for the Hamiltonian simulation within the HHL algorithm is based on data. For example, $A$ can contain the kernel matrix $K$ captured in the quantum system as a density matrix. Then, one needs to be able to efficiently compute $e^{-iK\Delta t}$, where $K$ is scaled by the trace of kernel matrix. Since $K$ is not sparse, the LMR protocol \citep{lloyd2014quantum} for the exponentiation of a non-sparse density matrix is used instead. The protocol involves simulation with a sparse operator $S$ to simulate the application of $e^{-iK\Delta t}$ to a state $\sigma$ in the following way:
\begin{equation}\label{eq:LMR}
\operatorname{Tr}_{1}\left\{e^{-i S \Delta t}(K \otimes \sigma) e^{i S \Delta t}\right\}=\sigma-i \Delta t[K, \sigma]+O\left(\Delta t^{2}\right)
\approx e^{-i K \Delta t} \sigma e^{i K \Delta t}.
\end{equation}
The operator  $S=\sum_{i,j}\ket{i}\bra{j}\otimes \ket{j}\bra{i}$ is the swap operator. The procedure relies on $\operatorname{Tr}_{1}\left\{S(K \otimes \sigma)\right\}=K\sigma$ and $\operatorname{Tr}_{1}\left\{(K \otimes \sigma)S\right\}=\sigma K$. The equation (\ref{eq:LMR}) summarizes the LMR protocol: approximating $e^{-i K \Delta t} \sigma e^{i K \Delta t}$ up to error $O(\Delta t^2)$ is equivalent to simulating a swap operator $S$, applying it to the state $K \otimes \sigma$ and discarding the first system by taking partial trace operation. Since the swap operator is sparse, its simulation is efficient. Therefore the LMR protocol provides an efficient way to approximate exponentiation of a non-sparse density matrix and can be readily used in HHL-based algorithms. 

\paragraph{Quantum LS-SVM.}
Quantum LS-SVM \citep{rebentrost2014quantum} uses partial trace to construct density operator corresponding to the kernel matrix $K$, as described above. Once the kernel matrix $K$ becomes available as a density operator, it proceeds by applying the HHL algorithm for solving the system of linear equations associated with LS-LSVM, using the LMR protocol described above for performing the density operator exponentiation $e^{-iK\Delta t}$.

\section{Quantum Semi-Supervised Least Square SVM}

We proposed here a quantum algorithm for solving Semi-Supervised Least- Squares SVM. As mentioned in Section \ref{secRKHS}, semi-supervised LS-SVM involves solving the following system of linear equations
\begin{align*}
 \left( \frac{1}{\gamma} K + K K  + \frac{1}{\gamma}   K L K \right) \vec{\alpha} =  K\vec{y}.
\end{align*}
In quantum setting the task is to generate $\ket{\boldsymbol{\alpha}}=\hat{A}^{-1}\ket{\textbf{Ky}}$, where the normalized $\hat{A}=\dfrac{A}{Tr(A)}$ for $A= \frac{1}{\gamma} K + K K  +\frac{1}{\gamma}   K L K$. The linear system differs from the one in LS-SVM, $( \gamma^{-1} K + K K) \alpha = Ky$ or simply $(K +\gamma^{-1}I)\alpha = y$. We have additional  $KLK$ term in the matrix to be inverted, and instead of the input vector of labels $\vec{y}$, we have the vector $K\vec{y}$. While this difference is of little significance for classical solvers, in quantum systems we cannot just multiply matrices and then apply quantum LS-SVM -- we are limited by the unitary nature of quantum transformations.

In order to obtain the solution to the quantum Semi-Supervised Least Square SVM, we will use the following steps. First, as shown in Section \ref{secQLap}, we will read in the graph information to obtain normalized graph Laplacian matrix in a quantum form as a density matrix $L$. We will also read in the kernel matrix as a density matrix, as described above. Given $K$, we will perform matrix multiplication $K\vec{y}$, using the variant of HHL and LMR protocol for matrix multiplication \citep{wiebe2012quantum} described above. Finally, as shown in Section \ref{sec:LMR}, we will use polynomial Hermitian exponentiation for performing the matrix inverse $(K+KK+K LK)^{-1}$, as described below.

\subsection{Quantum Input Model for the Graph Laplacian}
\label{secQLap}
Semi-supervised learning assumes that in addition to the training set, we have access to a graph $G$ connecting the $m$ training samples using $n$ edges. The connectivity structure of the graph captures the similarity of the samples, represented as vertices in the graph -- for example, each sample has an edge to $d$ samples most similar to it for some small $d$, often being chosen as a small fraction of the training size, $n$. More generally, the degrees of the vertices in the graph may differ, with each vertex $i$ having being connected to $d_i$ other vertices.
We assume that the graph is given as a graph adjacency lists data structure that can be accessed, in superposition, to create states corresponding to rows of the degree-normalized graph incidence matrix $G_I$, an $m \times n$, vertices-by-edges matrix. For an edge $e$ connecting vertices $i$ and $j$, with $i < j$, the incidence matrix will have entry $G_I[i,e]=-1/d_i$ and $G_I[j,e]=1/d_j$; we have $G_I[k,e]=0$ if edge $e$ does not involve vertex $k$. 

To have the graph available as a quantum density matrix for the quantum Semi-Supervised SVM, we observe that the normalized graph Laplacian $L$ is the Gram matrix of the rows of the $m \times n$ normalized graph incidence matrix $G_I$, that is, $L = G_I G_I^T$. Using oracle access to the graph, we can construct states $\ket{v_i}$ corresponding to rows of the graph incidence matrix $G_I$ for vertices $i$, $i=1,...,m$
\begin{align*}
\ket{v_i}=\frac{1}{\sqrt{d_i}}\sum_{e=1}^n G_I[i,e] \ket{e}.
\end{align*}
That is, state $\ket{v_i}$ has probability amplitude  $\frac{1}{\sqrt{d_i}}$ for each edge $e$, identified as $\ket{e}$, that is incident with vertex $i$, and null probability amplitude for all other edges in the graph. 

Based on the oracle that can construct individual rows of the incidence matrix, in superposition, we prepare a quantum state $\ket{G_I}$ that encodes all rows of the incidence matrix $G_I$ by preparing $\ket{v_i}$ states for all vertices in the graph
\begin{align*}
\ket{G_I}=\frac{1}{\sqrt{m}} \sum_{i=1}^m  \ket{i}\otimes \ket{v_i} =\frac{1}{\sqrt{m}} \sum_{i=1}^m  \frac{1}{\sqrt{d_i}}\sum_{t=1}^n G_I[i,t]    \ket{i} \otimes  \ket{t}.
\end{align*}
The normalized  graph Laplacian matrix $L = G_I G_I^T$, composed of inner products of the rows of $G_I$, is obtained in a form a density matrix by discarding the second part of the quantum system
\begin{equation}
\begin{split}
L &= Tr_2(\ket{G_I}\bra{G_I})= \frac{1}{m} \sum_{i,j=1}^m \ket{i}\bra{j}\otimes \frac{1}{\sqrt{d_i d_j}} \braket{v_i|v_j}=\frac{1}{m} \sum_{i,j=1}^m \frac{1}{\sqrt{d_i d_j}} \braket{v_i|v_j}\ket{i}\bra{j}.
\end{split}	
\end{equation}
The density matrix $L$, representing the Laplacian, will be used together with density matrix $K$, representing the kernel matrix, in the quantum method for training the Semi-Supervised SVM.

\subsection{Polynomial Hermitian Exponentiation for Semi-Supervised Learning}
\label{sec:LMR}
In quantum Semi-Supervised SVM, for computing the matrix inverse $(\gamma^{-1}K+KK+ \gamma^{-1}K LK)^{-1}$ via the HHL subroutine, we need to be able to efficiently compute $e^{-i (\gamma^{-1}K+KK+ \gamma^{-1}KLK) \Delta t} \sigma e^{i (\gamma^{-1}K+KK+ \gamma^{-1}KLK) \Delta t}$ given the ability to construct matrices $K$ and $L$ as quantum density matrices. For this purpose we adapt the Generalized LMR protocol for simulating Hermitian polynomials proposed in \citep{kimmel2017hamiltonian} to the specific case of Semi-Supervised SVM. For clarity, we first focus on simulation of $e^{-iKLK \Delta t}$; the other two terms, $K$ and $KK$, are also Hermitian polynomials and can be seen as special cases. Then, we discuss how the final dynamics involving the three-term sum $\gamma^{-1}K+KK+ \gamma^{-1}K LK$ can be obtained by creating a mixed state density corresponding to the sum of the three terms through sampling.

\paragraph{Simulating $\pmb{e^{iKLK\Delta t}}$.}
Let $D(\mathcal{H})$ denote the space of density operators associated with state space $\mathcal{H}$. Let $\rhoA, \rhoC, \rhoB \in D(\mathcal{H})$ be the density operators associated with the kernel matrix and the graph Laplacian, respectively. We will need two separate systems with the kernel matrix $K$; to distinguish between them we will denote the first as $\rhoA$ and the second as $\rhoC$; since $K$ is real and symmetric, these are indeed equal. The kernel and Laplacian matrices $K^\dagger, K, L$ are not sparse therefore we adapt the Generalized LMR technique for simulating Hermitian polynomials for the specific case $B=K^\dagger L K$. 

For adapting the Generalized LMR technique to our problem, we need to be able to generate a quantum state $\rho^\prime=\ket{0}\bra{0}\otimes \rhoY + \ket{1}\bra{1}\otimes \rhoZ$ with $Tr(\rhoY + \rhoZ)=1$, such that 
\begin{equation}\label{eq:G-LMR}
\begin{split}
\operatorname{Tr}_{1}\left\{\operatorname{Tr}_{3}\left\{e^{-iS^{\prime}\Delta}\left(\rho^{\prime}\otimes \sigma \right) e^{iS^{\prime}\Delta}\right\}\right\} = \sigma - i[B,\sigma]+ O(\Delta^2)
= 	e^{-iBt}\sigma e^{iBt} + O(\Delta^2),
\
\end{split}
\end{equation}
where $B=\rhoY - \rhoZ=\frac{1}{2}\rhoA\rhoB\rhoC+ \frac{1}{2}\rhoC\rhoB\rhoA = KLK$, and where $S^{\prime} := \ket{0}\bra{0}\otimes S+ \ket{1}\bra{1}\otimes (-S) $ is a controlled partial swap in the forward $(+S)$ and backward direction $(-S)$ in time. We have 
\[
e^{-iS^{\prime}\Delta} =  \ket{0}\bra{0}\otimes e^{-iS\Delta} +  \ket{1}\bra{1}\otimes e^{iS\Delta}.
\]
Therefore, with one copy of $\rho^{\prime}$, we obtain the simulation of $e^{-iB\Delta}$ up to error $O(\Delta^2)$. If we choose the time slice $\Delta= \delta/t$ and repeat the above procedure for $t^2/\delta$ times, we are able to simulate $e^{-iBt}$ up to error $O(\delta)$ using $n=O(t^2/\delta)$ copies of $\rho^{\prime}$.

\begin{figure}[h]
	\begin{center}
		\includegraphics[width=0.5\textwidth]{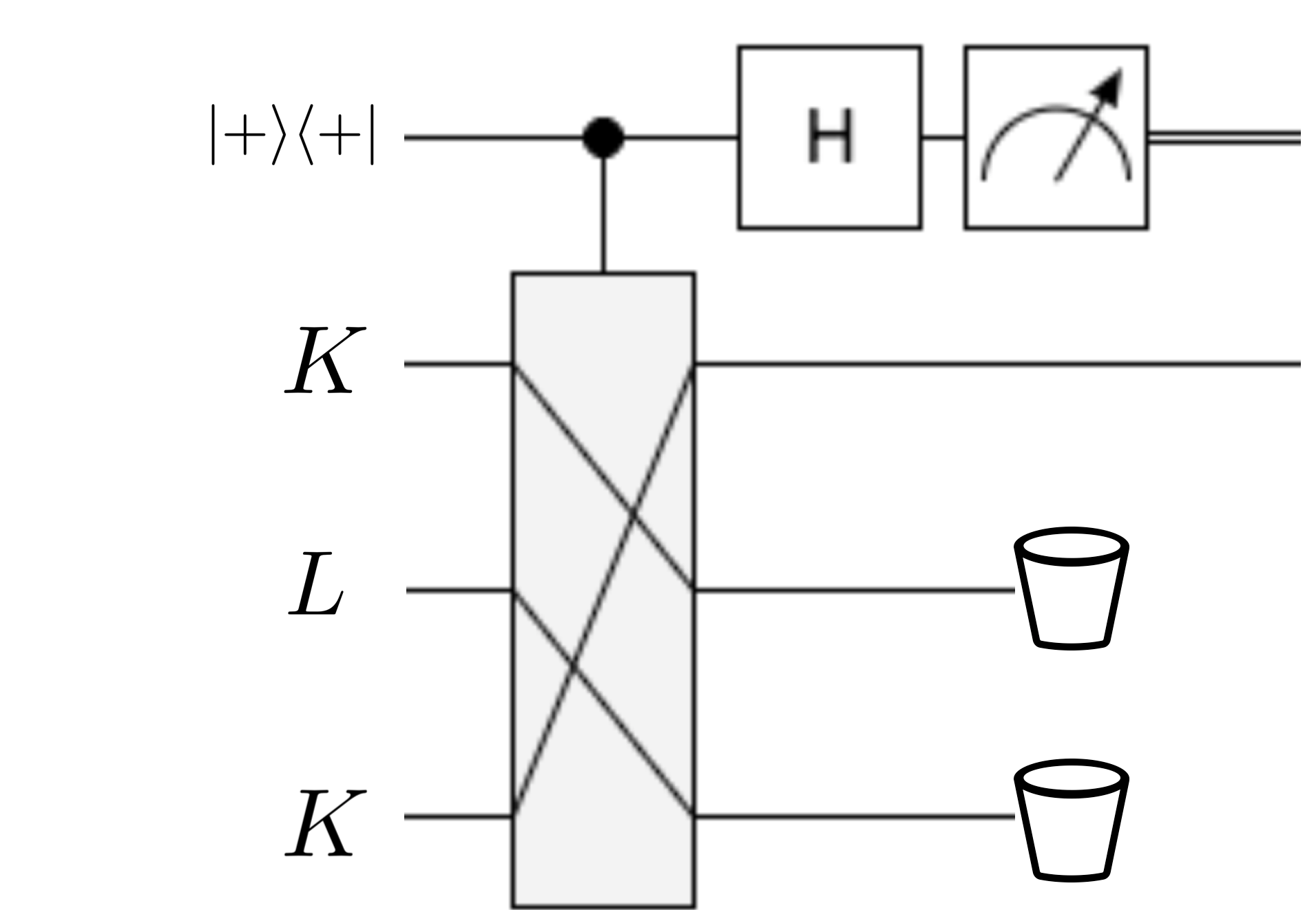}
	\end{center}
	\caption{\label{fig:subcicuit} Quantum circuit for creating $\rho^\prime=\ket{0}\bra{0}\otimes \rhoY+ \ket{1}\bra{1}\otimes \rhoZ$ by utilizing the Generalized LMR protocol \citep{kimmel2017hamiltonian} for density matrices representing the SVM kernel matrix $K$ and the graph Laplacian $L$. The circuit is to be read from left-to-right. Each wire at left shows its corresponding input state. The vertical rectangle denotes the  cyclic permutation operator $\operatorname{P}$ on $K,L,K$ defined in (\ref{eq:CP}). H is the Hadamard gate, and  the waste bins show partial trace. The measurement on the first quantum state is in computational basis. }
\end{figure}

\paragraph{Generating $\pmb{\rho^\prime=\ket{0}\bra{0}\otimes \rhoY + \ket{1}\bra{1}\otimes \rhoZ}$.} Figure \ref{fig:subcicuit} shows the quantum circuit for creating $\rho^\prime=\ket{0}\bra{0}\otimes \rhoY + \ket{1}\bra{1}\otimes \rhoZ$ such that $Tr(\rhoY + \rhoZ)=1$ and $B=\rhoY - \rhoZ=KLK$.   

The analysis of the steps preformed by the circuit depicted in Fig.\ref{fig:subcicuit} adapts the Generalized LMR protocol to the Semi-Supervised SVM problem. Let $\operatorname{P}$ be the cyclic permutation of three copies of $\mathcal{H_A}$ that operates as $\operatorname{P}\ket{j_1,j_2,j_3}=\ket{j_3,j_1,j_2}$.  In operator form it can be written as
\begin{equation}\label{eq:CP}
\operatorname{P} :=\sum_{j_{1}, j_{2}, j_{3}=1}^{\operatorname{dim} \mathcal{H}_{\mathrm{A}}}\left|j_{3}\right\rangle\left\langle j_{1}|\otimes| j_{1}\right\rangle\left\langle j_{2}|\otimes| j_{2}\right\rangle\left\langle j_{3}\right.|
\end{equation}
The input state to the circuit depicted in Fig. \ref{fig:subcicuit} is  
\begin{equation*}
\ket{+}\bra{+}\otimes\rhoA\otimes\rhoB\otimes\rhoC=\dfrac{1}{2}\sum_{i,j\in\{0,1\}}\ket{i}\bra{j}\otimes\rhoA\otimes\rhoB\otimes\rhoC.
\end{equation*}
Applying $\operatorname{P}$ on $\rhoA,\rhoB,\rhoC$ gives
\begin{equation*}
\begin{split}
\RNum{1}&= \frac{1}{2}[ \ket{0}\bra{0}\otimes\rhoA\otimes\rhoB\otimes\rhoC +\ket{0}\bra{1}\otimes\left(\rhoA\otimes\rhoB\otimes\rhoC\right)\operatorname{P}\\
&+\ket{1}\bra{0}\otimes \operatorname{P}\left(\rhoA\otimes\rhoB\otimes\rhoC\right)+\ket{1}\bra{1}\otimes \operatorname{P}\left(\rhoA\otimes\rhoB\otimes\rhoC\right)\operatorname{P}.
\end{split}
\end{equation*}
After discarding the third and second register sequentially by applying corresponding partial trace operators, we get
\begin{equation*}
\begin{split}
\RNum{2}=\operatorname{Tr}_{2}\left[\operatorname{Tr}_{3}(\RNum{1})\right] &= \ket{0}\bra{0}\otimes \frac{1}{2}\rhoA + \ket{0}\bra{1}\otimes \frac{1}{2} \rhoA\rhoB\rhoC \\
&+ \ket{1}\bra{0}\otimes \frac{1}{2} \rhoC\rhoB\rhoA + \ket{1}\bra{1}\otimes \frac{1}{2}\rhoC,
\end{split}
\end{equation*}
in this step $KLK$ term where the last line obtained from 
\begin{equation*}
\operatorname{Tr}_{2}\left[\operatorname{Tr}_{3}\left[\left(\rhoA\otimes\rhoB\otimes\rhoC\right)\operatorname{P}\right]\right]= \rhoA\rhoB\rhoC,
\end{equation*}
\begin{equation*}
\operatorname{Tr}_{2}\left[\operatorname{Tr}_{3}\left[\operatorname{P}(\rhoA\otimes\rhoB\otimes\rhoC)\right]\right]= \rhoC\rhoB\rhoA,
\end{equation*} 
\begin{equation*}
\operatorname{Tr}_{2}\left[\operatorname{Tr}_{3}\left[\operatorname{P}(\rhoA\otimes\rhoB\otimes\rhoC)\operatorname{P}\right]\right]= \rhoC.
\end{equation*}   
After applying a Hadamard gate $H=\frac{1}{\sqrt{2}}[(|0\rangle+|1\rangle)\langle 0|+(|0\rangle-|1\rangle)\langle 1|]$ on the first qubit of $\RNum{2}$, we get

\begin{equation*}
\begin{split}
\RNum{3} = H\otimes\mathds{1}(\RNum{2})H\otimes\mathds{1} &= \frac{1}{2}\left(\ket{0}\bra{0}+\ket{0}\bra{1}+\ket{1}\bra{0}+\ket{1}\bra{1} \right)\otimes \frac{1}{2} \rhoA \\
&+\frac{1}{2}\left( \ket{0}\bra{0}-\ket{0}\bra{1}+\ket{1}\bra{0}-\ket{1}\bra{1}\right)\otimes \frac{1}{2} \rhoA\rhoB\rhoC \\
&+\frac{1}{2}\left( \ket{0}\bra{0}+\ket{0}\bra{1}-\ket{1}\bra{0}-\ket{1}\bra{1}\right)\otimes\frac{1}{2} \rhoC\rhoB\rhoA \\
&+\frac{1}{2}\left( \ket{0}\bra{0}-\ket{0}\bra{1}-\ket{1}\bra{0}+\ket{1}\bra{1}\right)\otimes \frac{1}{2}\rhoC\\
&=\ket{0}\bra{0}\otimes\frac{1}{2}\left( \frac{1}{2}\rhoA+\frac{1}{2}\rhoA\rhoB\rhoC + \frac{1}{2}\rhoC\rhoB\rhoA +\frac{1}{2}\rhoC \right)\\
&+\ket{0}\bra{1}\otimes\frac{1}{2}\left( \frac{1}{2}\rhoA-\frac{1}{2}\rhoA\rhoB\rhoC + \frac{1}{2}\rhoC\rhoB\rhoA -\frac{1}{2}\rhoC \right)\\
&+\ket{1}\bra{0}\otimes\frac{1}{2}\left( \frac{1}{2}\rhoA+\frac{1}{2}\rhoA\rhoB\rhoC - \frac{1}{2}\rhoC\rhoB\rhoA -\frac{1}{2}\rhoC \right)\\
&+\ket{1}\bra{1}\otimes\frac{1}{2}\left( \frac{1}{2}\rhoA-\frac{1}{2}\rhoA\rhoB\rhoC - \frac{1}{2}\rhoC\rhoB\rhoA +\frac{1}{2}\rhoC \right).
\end{split}
\end{equation*} 
The last step is applying a measurement in computational basis $\{\ket{0}\bra{0},\ket{1}\bra{1}\}$ on the first register to obtain our desired state $\rho^\prime$,
\begin{multline*}
\begin{split}
	\RNum{4} &=\ket{0}\bra{0}\otimes\frac{1}{2}\left( \frac{1}{2}\rhoA+\frac{1}{2}\rhoA\rhoB\rhoC + \frac{1}{2}\rhoC\rhoB\rhoA +\frac{1}{2}\rhoC \right)\\
&+\ket{1}\bra{1}\otimes\frac{1}{2}\left( \frac{1}{2}\rhoA-\frac{1}{2}\rhoA\rhoB\rhoC - \frac{1}{2}\rhoC\rhoB\rhoA +\frac{1}{2}\rhoC \right)
\end{split}
\end{multline*}
We can see that by defining $\rhoY=\frac{1}{2}\left( \frac{1}{2}\rhoA+\frac{1}{2}\rhoA\rhoB\rhoC + \frac{1}{2}\rhoC\rhoB\rhoA +\frac{1}{2}\rhoC \right) $ and 
$\rhoZ=\frac{1}{2}\left( \frac{1}{2}\rhoA-\frac{1}{2}\rhoA\rhoB\rhoC - \frac{1}{2}\rhoC\rhoB\rhoA +\frac{1}{2}\rhoC \right)$ the final state is in the form of $\rho^{\prime}=\ket{0}\bra{0}\otimes \rhoY+ \ket{1}\bra{1}\otimes \rhoZ$ where $Tr(\rhoY + \rhoZ)=1$, and we obtain
$
\rhoY - \rhoZ = \frac{1}{2}\rhoA\rhoB\rhoC + \frac{1}{2}\rhoC\rhoB\rhoA = B$.

Now with having the output state $\rho^{\prime}$ we are ready to apply the generalized LMR in (\ref{eq:G-LMR}) to simulate $e^{-iKLK\Delta t}\sigma e^{iKLK\Delta t}$ up to error $O(\Delta^2)$. Comparing the LMR technique in equation (\ref{eq:LMR}) with the generalized LMR for the special case of $KLK$ in equation (\ref{eq:G-LMR}), we see approximating $e^{-i KLK \Delta t} \sigma e^{i KLK \Delta t}$ up to error $O(\Delta t^2)$ is equivalent to simulating the controlled partial swap operator $S^\prime$, applying it to the state $\rho^\prime \otimes \sigma$ and discarding the third and first systems by taking partial trace operations, respectively. Since $S^\prime$ is also sparse, and its simulation is efficient, the generalized LMR technique offers an efficient approach for simulating $e^{i KLK \Delta t}$.

\paragraph{Simulating joint evolution involving $K$, $KK$, and $KLK$.}

Simulation involving a sum  $\gamma^{-1}K+KK+\gamma^{-1}KLK$ can be obtained, in the generalized LMR protocol, by creating a mixed state $\rho^\prime_{joint}=\ket{0}\bra{0}\otimes \rhoY_{joint} + \ket{1}\bra{1}\otimes \rhoZ_{joint}$ such that $ \rhoY_{joint} - \rhoZ_{joint} = \gamma^{-1}K+KK+\gamma^{-1}KLK$, with equality up to normalization of the term coefficients. As describe above for the case of $KLK$ term alone, ability to construct copies of this state translates to the ability of simulating the Hermite polynomial. 

To extend the process beyond just the $KLK$ term, in addition to $\rhoY_{KLK}=\rhoY$ and $\rhoZ_{KLK}=\rhoZ$ leading to $\rho^\prime_{KLK}=\ket{0}\bra{0}\otimes \rhoY_{KLK} + \ket{1}\bra{1}\otimes \rhoZ_{KLK}$ construction shown above, we need to be able to construct copies of the state $\rho^\prime_{KK}=\ket{0}\bra{0}\otimes \rhoY_{KK} + \ket{1}\bra{1}\otimes \rhoZ_{KK}$ involving pairs of densities $\rhoY_{KK}$, $\rhoZ_{KK}$, and copies of  $\rho^\prime_{K}=\ket{0}\bra{0}\otimes \rhoY_{K} + \ket{1}\bra{1}\otimes \rhoZ_{K}$ involving $\rhoY_{K}$, $\rhoZ_{K}$; these states can be constructed in a similar fashion as described above for $KLK$. 

In simulating the evolution $e^{-i (\gamma^{-1}K+KK+ \gamma^{-1}KLK) \Delta t} \sigma e^{i (\gamma^{-1}K+KK+ \gamma^{-1}KLK) \Delta t}$, which involves constructing and using multiple copies of state $\rho^\prime_{joint}$, the generalized LMR protocol uses individual states corresponding $\rho^\prime_{K}$, $\rho^\prime_{KK}$, and $\rho^\prime_{KLK}$ with probabilities given by the coefficients of the three terms, $\frac{1}{\gamma}$, 1, $\frac{1}{\gamma}$, respectively. This probabilistic process is then equivalent to using a single mixed state $\rho^\prime_{joint}$.

\subsection{Classification using a Trained Quantum Model}
\label{secClassif}
 The result of the procedure for training a quantum semi-supervised LS-SVM is a quantum state
\begin{equation}
	\ket{\vec{\alpha}} =  \frac{1}{\sqrt{ \sum_{j=1}^m \alpha_j ^2}} \left( \sum_{j=1}^m \alpha_j \ket{j}\right),
\end{equation}
representing the normalized solution vector $\alpha$. Unlike in classical semi-supervised kernel SVM, $\alpha$ is not readily accessible. 
In the quantum setting, we can follow the swap test-based approache proposed for quantum LS-SVM \citep{rebentrost2014quantum}. We first construct a query state based on new data point $\ket{\vec{x_{new}}}$ 
\begin{equation}
	\ket{\vec{x_q}} = \frac{1}{\sqrt{m\norm{\ket{\vec{x_{new}}}}^2 +1}}\left(\sum_{j=1}^{m}\ket{j} \otimes \norm{\ket{\vec{x_{new}}}} \ket{\vec{x_{new}}}\right).
\end{equation}
Then, based on the ability to construct $\ket{\vec{\alpha}}$ and oracle access to the training data, we construct the following state
\begin{equation}
	\ket{\vec{s}} = \frac{1}{\sqrt{ \sum_{j=1}^m \alpha_j^2 \norm{\vec{x_j}}^2}}   \left(   \sum_{j=1}^{m} \alpha_j \ket{j} \otimes \norm{\ket{\vec{x_j}}} \ket{\vec{x_j}} \right).
\end{equation}
Similarly to classical setting, where class prediction is performed via inner products between the new samples and the training samples, 
we can utilize sign of the inner product the states $\ket{\vec{x_q}}$ and $\ket{\vec{s}}$ for obtaining the class of the new data point $\vec{x_{new}}$
\begin{equation} \label{classification}
	y_{\vec{x_{new}}} = \mathrm{sign}(\braket{\vec{x_q}|\vec{s}}) = \mathrm{sign} \left(C \left(   \sum_{j=1}^m \alpha_j \norm{\ket{\vec{x_{new}}}} \norm{\ket{\vec{x_j}}} \braket{\vec{x_{new}}|\vec{x_j}}\right)\right),
\end{equation}
where $ C= 1/{\sqrt{(m\norm{\ket{\vec{x_{new}}}}^2 +1)(\sum_{j=1}^m \alpha_j^2 \norm{\vec{x_j}}^2)}}$.
For implementing (\ref{classification}) on a quantum computer, we perform a procedure similar to swap test \citep{rebentrost2014quantum}. Given two quantum states $\ket{\psi}$ and $\ket{\phi}$, the swap test is a quantum operation that measures the overlap between the two states. For this aim, using an ancilla qubit, we construct $\ket{\psi} = \frac{1}{\sqrt{2}} \left( \ket{0} \ket{\vec{s}} + \ket{1} \ket{\vec{x_q}} \right)$. Next, we apply Hadamard gate on the  first qubit (the ancilla) of $\ket{\psi}$. We repeat the process multiple times and quantify the probability of successfully measuring the ancilla in the state $\ket{-}= \frac{1}{\sqrt{2}}(\ket{0}-\ket{1})$; the success probability is $P = \dfrac{1}{2}(1-\braket{\vec{x_q}|\vec{s}})$ and can be approximated up to $\epsilon$ error with $O(P(1-P)/\epsilon^2)$ tries. Once $P$ is estimated, it allows for predicting the class, since $P=\frac{1}{2}$ is the threshold separating positive from negative prediction.    

\begin{algorithm}[h]
	\renewcommand{\algorithmicrequire}{\textbf{Input:}}
	\renewcommand{\algorithmicensure}{\textbf{Output:}}
	\caption{Quantum Semi-Supervised LS-SVM}
	\label{QSLS_SVM}
	\begin{algorithmic}[1]
		\Require The datapoint set $\{x_1,...x_l,...x_m\}$ with the first $l$ data points labeled and the rest unlabeled, $\textbf{y}=(y_1,...,y_l) $ and the graph $G$ 
		\Ensure  The classifier $\ket{\alpha}=A^{-1}\ket{y}$
		\State \textbf{Quantum data preparation.} Encode classical data points into quantum data points using quantum oracles $O_x:\{x_1,...x_l,...x_m\} \mapsto \ket{X}=\frac{1}{\sqrt{\sum_{i=1}^m \norm{x_i}^2}} \sum_{i=1}^m\ket{i}\otimes \norm{x_i}\ket{x_i}$ and $O_x:\textbf{y}\mapsto \ket{y}$.
		\State \textbf{Quantum Laplacian preparation.} Prepare quantum density matrix using oracle access to $G$ (Section \ref{secQLap}).
		\State \textbf{Matrix multiplication.} Compute the matrix multiplication $\ket{Ky}=K\ket{y}$ via \citep{wiebe2012quantum}.
		\State \textbf{Matrix inversion.} Compute the matrix inversion $\ket{\alpha,}=A^{-1}\ket{\kappa}$ via HHL algorithm.
		A quantum circuit for the HHL algorithm has three main steps:
		\State\hspace{\algorithmicindent}\emph{Phase estimation}, including efficient Hamiltonian simulation (Section \ref{sec:LMR})
		\State\hspace{\algorithmicindent}\emph{Controlled rotation}
		\State\hspace{\algorithmicindent}\emph{Uncomputing} 
		\State \textbf{Classification.} Based on \textbf{Swap test} algorithm, same as in Quantum LS-SVM (Section \ref{secClassif}).
	\end{algorithmic}
\end{algorithm}
\section{Complexity of Quantum Semi-Supervised LS-SVM }

The complexity of the algorithm is $O(\kappa^{3} \varepsilon^{-3} \log mp)$, where $\varepsilon$ is the desired error and $\kappa=1/\sigma$ is the effective conditioning number, defined by filtering out eigenvalues of the matrix to be inverted that are below $\sigma$. The quantum method offers exponential speedup  over the classical time complexity for solving SVM as a quadratic problem, which requires time $O(\log(\epsilon^{-1})poly(p,m))$, where $\epsilon$ is the desired error. The exponential speedup in $p$ occurs as the result of fast quantum computing of kernel matrix, and relies on the existence of efficient oracle access to data. The speedup on $m$ is due to applying quantum matrix inversion for solving LS-SVM, which is inherently due to fast algorithm for exponentiation of a resulting non-sparse matrix. However, to achieve the exponential speedup in $m$, the matrix to be inverted has to be very low-rank; for more typical low- to medium-rank matrices, the speedup is polynomial in $m$. Compared to quantum LS-SVM \citep{rebentrost2014quantum}, our algorithm introduces three additional steps: preparing the Laplacian density matrix, quantum matrix multiplication $Ky$, and Hamiltonian simulation for $\gamma^{-1}K + KK + \gamma^{-1}K LK$ instead of just $K$. The first step involves oracle access to a sparse graph adjacency list representation, which is at least as efficient as the oracle access to non-sparse data points. The matrix multiplication step is a variant of the HHL and LMR protocol and is dominated in term of complexity by the Hamiltionian simulation step. The Hamiltonian simulation involving HHL and generalized LMR protocol involves simulating a sparse conditional partial swap operator, which results an efficient strategy for simulating Hermite polynomials in time $\tilde{O}(\log(m) \Delta t)$, where the notation $\tilde{O}$ hides more slowly growing factors in the Hamiltonian simulation \citep{berry2007efficient}.   

The quantum Semi-Supervised SVM, similarly to other quantum machine learning methods, relies on  a strong assumptions about the input-output model: the existence of an efficient approach for preparing input data (mainly via QRAM), and learning partially from the quantum output state (via quantum measurement). 
Recently, the consequences of adopting similar input-output assumptions for classical algorithms have been explored under the umbrella of quantum-inspired algorithms \citep{tang2019quantum}, also known as a dequantized algorithms. These algorithms solve a classical equivalent of a quantum machine learning problem in a setting where the data input/output model is designed to mimic the assumptions underlying QRAM and quantum measurements. Specifically, analogous to using QRAM input model and quantum measurement, these classical algorithms exploit \emph{$\ell^2$-norm sampling and query access}, also known as \emph{importance sampling} or \emph{length-square sampling} in randomized linear algebra literature. 
These assumptions make the comparison between quantum algorithms and their classical, dequantized counterparts more fair.

 Analogous to the assumption one can efficiently prepare a quantum state $\ket{x}$ proportional (up to normalization) to some input vector $x$, a quantum-inspired algorithm is assisted by an input model called sampling and query access. One has sampling and query access to a vector $x\in \complex^n$ if the following queries can be done in $\BigPolyOh{1}$:
 \begin{enumerate}[label=(\alph*)]
 	\item given an index $i\in [n]$, output the $i$th element $x_i$,
 	\item sample an index $j \in [n]$ with probability $\frac{|x_j^2|}{\norm{x}^2}$, 
 	\item output the $\ell^2$-norm $\norm{x}$.
 \end{enumerate}
  For a matrix $A\in \complex^{m\times n}$ the model gives the sampling and query access to each row $A(i,.)$ and access to a vector with elements corresponding to $\ell^2$-norm of $A$'s rows.\\
 The fundamental difference between quantum-inspired algorithms and traditional classical algorithms is that via importance sampling, their runtime is independent of the dimension of input data, and thus it builds a setting comparable with quantum machine learning algorithms aided by QRAM. Recently \citep{chia2020sampling} introduced an algorithmic framework for quantum-inspired classical algorithms on low-rank matrices that generalizes a series of previous work, recovers existing quantum-inspired algorithms such as quantum-inspired recommendation systems \citep{tang2019quantum}, quantum-inspired principal component analysis \citep{tang2018quantum}, quantum-inspired low-rank matrix inversion \citep{gilyen2018quantum}, and quantum-inspired support vector machine \citep{ding2019quantum}. 
 
 It is natural to ask how our proposed quantum algorithm's complexity compares to the quantum-inspired classical setting. Quantum Semi-supervised SVM has complexity
 \begin{align*}
 \BigPolyOh{\sigma^{-3} \varepsilon^{-3} \log mp }
  \end{align*} 
The value of $\sigma$ is a crucial parameter that may increase the complexity above logarithmic factors in $m$ and $p$. It is user controlled, and denotes a constant such that that the magnitudes of $\hat{F}$'s eigenvalues $\lambda_i$ that are used in the computations satisfy $\sigma \leq|\lambda_i|$, and smaller eigenvalues are ignored. Here, matrix $\hat{F}= {F}/{tr(F)}$ is a normalized matrix that needs to be inverted to find the solution to the SVM problem.

Quantum-inspired SVM \citep{ding2019quantum,chia2020sampling} for the same $\hat{F}$ with trace, operator norm, and Froebenius norm bounded by 1, and with threshold $\sigma$ on the singular values retained in the computation, has complexity 
\begin{align*}
\widetilde{O}\left(\frac{\|\hat{F}\|_{\mathrm{F}}^{6}\|\hat{F}\|^{22}}{\sigma^{28} \varepsilon^{6}} \eta^{6} \log ^{3} \frac{1}{\delta}\right),
\end{align*}
where $ \varepsilon$ is the error, as in Quantum SVM. Parameters $\eta$ and $\delta$ have no corresponding meaning in quantum SVM; without loss of generality we consider their values as a constant. 

To illustrate the differences in complexity of the quantum and dequantized algorithms, we construct a family of examples where the input $m \times m$ matrix $F$ is a diagonal matrix with $q$ ones and $m-q$ zeros on the diagonal. The trace of $F$ is $q$, and thus $\hat{F}$ is a diagonal matrix with value $1/q$ and null on the diagonal. The operator norm $\norm{\hat{F}}=\lambda_{max} = 1/q$, while the Froebenius norm $\norm{\hat{F}}_F=\sqrt{\sum_{i} \lambda_i^2} = \sqrt{q \frac{1}{q}^2}=1/\sqrt{q}$. Since all nonzero singular values are equal to $1/q$, setting $\sigma = 1/q$ is the only reasonable choice. 
For a given $q$, the complexity of the quantum algorithm is 
\begin{align*}
\BigPolyOh{q^3 \varepsilon^{-3} \log mp },
\end{align*} 
while the complexity of the dequantized algorithm is
\begin{align*}
\BigPolyOh{
		\frac{   (1/q)^{6} (1/\sqrt{q})^{22}}{(1/q)^{28}  \varepsilon^{-6}} \eta^{6} \log ^{3} \frac{1}{\delta}
} = \BigPolyOh{ q^{-6 -11 +28} \varepsilon^{-6} \eta^{6} \log ^{3} \frac{1}{\delta} }= \BigPolyOh{ q^9 \varepsilon^{-6}  \eta^{6} \log ^{3} \frac{1}{\delta} 
},
\end{align*}

In the full-rank case, $q=m$, we have $
\BigPolyOh{ m^3 \varepsilon^{-3} \log mp }$ complexity for the quantum method, and $\BigPolyOh{ m^9 \varepsilon^{-6}  \eta^{6} \log ^{3} \frac{1}{\delta} 
}$ for the dequantized algorithm; both are polynomial in terms of $m$, and do not offer advantage in terms of $m$ over classical methods involving matrix inversion. Such a full-rank case is unlikely to appear in practice in Semi-Supervised SVMs, where 

In the extremely low-rank case of $q=1$, the complexity of both methods is similar, we have $
\BigPolyOh{ \varepsilon^{-3} \log mp }$ complexity for the quantum method, and $\BigPolyOh{  \varepsilon^{-6}  \eta^{6} \log ^{3} \frac{1}{\delta} 
}$ for the dequantized algorithm; both are sublinear in terms of $m$ and $p$; the quantum method has better dependency on the error $\varepsilon$ but includes logarithmic factor $\log mp$ that is absent in the dequantized method.

Consider also a more realistic low-rank case where the rank slowly grows with $m$, for example $q=\sqrt[6]{m}$. In this case, we have $
\BigPolyOh{ \sqrt{m} \varepsilon^{-3} \log mp }$ complexity for the quantum method, and $\BigPolyOh{ m^{3/2} \varepsilon^{-6}  \eta^{6} \log ^{3} \frac{1}{\delta} 
}$ for the dequantized algorithm. The quantum algorithm is sublinear, while the dequantized one is not. 
Based on the distribution of eigenvalues of the Laplacian for real-world graphs \citep{eikmeier2017revisiting} and for kernel matrices \citep{wathen2015spectral}, which decay fast, we can expect this is the most realistic case. 

These observations agree with a recent work \citep{arrazola2019quantum} that studied the performance of  quantum-inspired algorithms in practice and concluded that their performance degrades significantly when the rank and condition number of the input matrix are increased, and high performance requires very low rank and condition number. 

\section{Conclusions}

Quantum Semi-Supervised SVM uses unlabeled training samples to gain more information about the underlying distribution over the sample input space, which helps improve the decision boundary between the classes compared to only using labeled samples. In classical algorithms, the Laplacian of the graph that connects labeled and unlabeled samples can be incorporated into the SVM objective function in a straightforward way. However, in the quantum SVM, where the kernel matrix is represented by a density matrix and Hamiltonian simulation is used for minimizing the objective function, incorporating the graph Laplacian, also represented as a density matrix, becomes a challenge. We address this difficulty by adapting the Generalized LMR protocol to the Semi-Supervised SVM objective function, composed of the quadratic terms that involve the kernel and graph Laplacian matrices. We compared the complexity of the resulting algorithm with the complexity of classical methods, as well as with recently proposed dequantized algorithms, and showed that in the most realistic, low-rank case, the proposed quantum method achieves sublinear complexity, providing polynomial speedup over classical solvers.

\section*{Acknowledgments}
T.A. is supported by NSF grant IIS-1453658.

\bibliographystyle{unsrt}

\newcommand{\noopsort}[1]{}

\end{document}